%% file: main.tex
\documentclass[conference]{IEEEtran}
\IEEEoverridecommandlockouts
\usepackage{cite}
\usepackage{amsmath,amssymb,amsfonts}
\usepackage{algorithmic}
\usepackage{graphicx}
\usepackage{textcomp}
\usepackage{xcolor}
\usepackage{comment}
\usepackage{subcaption}

\def\BibTeX{{\rm B\kern-.05em{\sc i\kern-.025em b}\kern-.08em
    T\kern-.1667em\lower.7ex\hbox{E}\kern-.125emX}}

\usepackage{eso-pic}
\newcommand{\mycopyrighttext}{%
  \footnotesize
  \noindent
  \textcopyright~2025 IEEE. Personal use of this material is permitted. Permission from IEEE must be obtained for all other uses, in any current or future media, including reprinting/republishing this material for advertising or promotional purposes, creating new collective works, for resale or redistribution to servers or lists, or reuse of any copyrighted component of this work in other works.\\
 IEEE International Automated Vehicle Validation Conference (IAVVC), September 30 - October 2, 2025.
}

\AtBeginDocument{%
  \AddToShipoutPicture*{%
    \AtPageUpperLeft{%
      \put(55, -40){% move right 2pt, down 40pt from upper-left corner
        \parbox{\dimexpr\textwidth-4pt\relax}{\raggedright \mycopyrighttext}
      }%
    }
  }
}
\title{Bridging Simulation and Usability: A User-Friendly Framework for Scenario Generation in CARLA
}
\author{Ahmed Abouelazm\textsuperscript{\textasteriskcentered}$^{1}$, Mohammad Mahmoud\textsuperscript{\textasteriskcentered}$^{2}$, Conrad Walter\textsuperscript{\textasteriskcentered}$^{2}$, Oleksandr Shchetsura$^{2}$, \\ Erne Hussong$^{2}$, Helen Gremmelmaier$^{1}$, J. Marius Zöllner$^{1,2}$
\thanks{\textasteriskcentered~These authors contributed equally to this work}%
\thanks{$^{1}$Authors are with the FZI Research Center for Information Technology, Germany {\tt\small abouelazm@fzi.de}}%
\thanks{$^{2}$Authors are with the Karlsruhe Institute of Technology (KIT), Germany}%
}

\begin{document}
\maketitle
\thispagestyle{empty}
\pagestyle{empty}

\begin{abstract}
\input{sections/0_abstract}
\end{abstract}


\input{sections/1_introduction}
\input{sections/2_related_work}
\input{sections/3_Methodology}
% \input{sections/4_setting}
\input{sections/5_evaluation}
\input{sections/6_conclusion}

\section*{ACKNOWLEDGMENT}
The research leading to these results is funded by the German Federal Ministry for Economic Affairs and Energy within the project “Safe AI Engineering – Sicherheitsargumentation befähigendes AI Engineering über den gesamten Lebenszyklus einer KI-Funktion". The authors would like to thank the consortium for the successful cooperation.

{
    \bibliographystyle{IEEEtran}
    \bibliography{references}
}

\end{document}

%% file: sections/0_abstract.tex
Autonomous driving promises safer roads, reduced congestion, and improved mobility, yet validating these systems across diverse conditions remains a major challenge. Real-world testing is expensive, time-consuming, and sometimes unsafe, making large-scale validation impractical. In contrast, simulation environments offer a scalable and cost-effective alternative for rigorous verification and validation. A critical component of the validation process is scenario generation, which involves designing and configuring traffic scenarios to evaluate autonomous systems' responses to various events and uncertainties. However, existing scenario generation tools often require programming knowledge, limiting accessibility for non-technical users. To address this limitation, we present an interactive, no-code framework for scenario generation. Our framework features a graphical interface that enables users to create, modify, save, load, and execute scenarios without needing coding expertise or detailed simulation knowledge. Unlike script-based tools such as Scenic or ScenarioRunner, our approach lowers the barrier to entry and supports a broader user base. Central to our framework is a graph-based scenario representation that facilitates structured management, supports both manual and automated generation, and enables integration with deep learning-based scenario and behavior generation methods. In automated mode, the framework can randomly sample parameters such as actor types, behaviors, and environmental conditions, allowing the generation of diverse and realistic test datasets. By simplifying the scenario generation process, this framework supports more efficient testing workflows and increases the accessibility of simulation-based validation for researchers, engineers, and policymakers. Future extensions will explore integrating real-world traffic data and incorporating heuristic-guided methods to improve scenario diversity, realism, and coverage, ultimately supporting more robust and reliable autonomous vehicle deployment.

%% file: sections/1_introduction.tex
\section{Introduction}
\label{sec:Introduction}

%\textbf{What does autonomous driving promise?} (Erne) Autonomous driving aims to improve road safety, reduce congestion, and improve mobility for those with limited access to traditional vehicles. Since human error causes 94\% of accidents \cite{nhtsa2015crashes}, autonomous vehicles (AVs) could significantly lower crash rates \cite{anderson2016autonomous}. Additionally, optimized traffic flow can ease congestion and reduce fuel consumption \cite{litman2022av}, while AVs improve mobility for older adults and individuals with disabilities, promoting social inclusion \cite{fagnant2014shared}. 
Autonomous driving (AD) holds the promise of transforming transportation by enhancing road safety, reducing traffic congestion, and expanding mobility for underserved populations \cite{liu2024investigating}. With human error responsible for up to 90\% of traffic accidents, autonomous vehicles (AVs) have the potential to significantly lower crash rates by minimizing driver-related risks \cite{abdel2024matched}. In addition, AVs could potentially reduce traffic congestion and fuel consumption by up to 20\% \cite{othman2022exploring}, thanks to more efficient driving behaviors and optimized traffic flow \cite{litman2022av}. Moreover, AVs offer new opportunities for increased independence and accessibility, particularly for senior citizens and individuals with disabilities, thereby fostering greater social inclusion and mobility equity \cite{golbabaei2024enabling}.

%\textbf{Importance of validation and verification of autonomous driving systems.} (Erne) Before autonomous driving can deliver on its promises, its safety and reliability must be guaranteed. Given the inherent complexity of AVs, this requires rigorous validation and verification across a wide range of conditions. \cite{rajabli2020software} \cite{koopman2016challenges}

Before AD can fully realize its transformative potential, ensuring the safety and reliability of AVs is paramount. Due to the inherent complexity of AV systems, which integrate advanced perception, decision-making, and control components, achieving this goal demands comprehensive validation and verification across diverse operating scenarios and edge cases \cite{rajabli2020software,koopman2016challenges}. % equivalent to over 11 million hours at average urban speeds 

Validation of AVs in real-world environments is largely impractical, primarily due to their nature as safety-critical systems. In particular, large-scale testing on public roads raises significant ethical, legal, and societal concerns, particularly given the potential consequences of system failures involving human lives \cite{koopman2016challenges, leahy2024grandchallengesverificationautonomous}. Moreover, real-world validation is inherently inefficient—studies estimate that demonstrating, with 95\% confidence, that an autonomous vehicle is just 20\% safer than a human driver would require approximately 275 million miles of incident-free driving \cite{kalra2016driving}. These limitations highlight the necessity for safe and scalable validation methodologies that ensure system safety without compromising ethical or legal responsibilities.

%\textbf{Challenges against validation in the real world.} (Sascha) However, validation in the real world is challenging mainly due to uncertainty and context. Uncertainty arises from unknown variables in dynamic settings, making it difficult to anticipate and model all possible scenarios. Context challenges include legal, ethical, and social responsibilities. \cite{leahy2024grandchallengesverificationautonomous} %\textbf{Simulation as a valid alternative.} (Sascha) To address these challenges, simulation offers a powerful and cost-effective alternative for validating autonomous vehicles by enabling extensive testing and iterative refinement in a virtual environment before real-world deployment. \cite{10242366} 

To address these challenges, simulation environments have become an essential tool for the validation and verification of autonomous vehicles, offering a safe, scalable, and cost-efficient alternative to evaluate system behavior across a wide range of driving scenarios \cite{hu2023simulation}. By enabling extensive testing and iterative development in a virtual environment, simulation allows developers to explore edge cases, optimize performance, and identify potential failures—all without the risks and constraints of real-world deployment \cite{goyal2025system}.

%\textbf{Scenario generation and its importance to the validation process.}(Conrad)Scenario generation is a key aspect of this validation process, as it enables the systematic creation of diverse and challenging test cases. It helps assess the autonomous system's performance under various conditions, addressing the limitations of real-world testing by allowing the creation of rare but critical situations as well as common ones.

Scenario generation plays a critical role in the simulation-based validation of autonomous vehicles, enabling the systematic construction of diverse, representative, and challenging test cases \cite{bogdoll2022one}. By allowing controlled variation of environmental, traffic, and behavioral parameters, scenario generation facilitates the evaluation of system behavior across both frequently encountered conditions and rare, safety-critical edge cases \cite{Kurz2023ScenarioValidation}. This approach addresses key limitations of real-world testing by making it possible to replicate hazardous situations without endangering human lives or relying on chance encounters.

%\textbf{Research gap in scenario generation frameworks currently available.} (Conrad) Widely available scenario generation frameworks require varying levels of expertise (see Section II); however, no solution currently enables domain experts to intuitively create and execute scenarios without prior expertise. This gap limits researchers, policymakers, and industry professionals from effectively leveraging their knowledge and experience to contribute to scenario generation, and thus to the validation of autonomous vehicles.

\textbf{Research Gap. }Although scenario generation frameworks such as OpenSCENARIO \cite{openscenario2020} and Scenic \cite{fremont2019scenic} are publicly available, they typically require varying degrees of technical expertise to be used effectively. At present, no publicly available framework enables domain experts to intuitively define and execute test scenarios without prior experience in simulation technologies or programming. This limitation restricts the integration of valuable domain-specific knowledge into the scenario generation process and, by extension, into the broader validation pipeline for autonomous vehicles. Addressing this gap is crucial for promoting interdisciplinary engagement and ensuring that scenario design reflects both technical requirements and real-world operational concerns.

% \textbf{Brief summary of our proposed framework and how it addresses the research gap.} (Mo) To address these limitations, we present an interactive, user-friendly framework for scenario creation in CARLA. Designed with accessibility in mind, it allows users to intuitively create, modify, and execute traffic scenarios without requiring programming skills or prior familiarity with CARLA’s architecture. This significantly lowers the entry barrier to scenario-based testing. By combining ease of use with a fully integrated workflow, the framework directly addresses the research gap in accessible and integrated scenario generation tools.

\textbf{Contribution. }To address this research gap, we propose an interactive and user-friendly framework for scenario generation in CARLA \cite{dosovitskiy2017carla}, an open-source simulator widely adopted for AD research. The proposed framework is designed with accessibility in mind, allowing users to intuitively define, modify, and execute traffic scenarios without requiring any programming expertise. By removing the technical barrier, the tool enables domain experts, such as policymakers, engineers, and safety analysts, to actively contribute to scenario development and validation. A central component of our contribution is a graph representation of scenarios, which facilitates the automatic generation of diverse scenario datasets for validation purposes and supports seamless integration with deep learning frameworks for data-driven model development and evaluation.

%% file: sections/2_related_work.tex
\section{Related Work}
In this section, we provide an overview of simulation environments commonly employed in AD. We then examine widely used scenario generation frameworks, highlighting their capabilities and limitations. Lastly, we review recent advances in data-driven scenario generation techniques.

\subsection{Simulation Environments}
Simulation environments for AD play a crucial role in the development, testing, and validation of safety-critical systems. Lightweight simulators such as SUMO \cite{krajzewicz2002sumo}, MetaDrive \cite{li2022metadrive}, and NuPlan \cite{caesar2021nuplan} focus primarily on modeling traffic dynamics and the interaction between agents. Their strength lies in efficiently simulating large-scale environments and realistic driving behaviors, making them well-suited for high-level planning and decision-making tasks. However, these platforms typically operate with simplified representations of the world and lack the capability to generate high-quality, physics-based sensor data. This limitation makes them less suitable for the validation of perception and sensor fusion components.

Conversely, many state-of-the-art simulators are built on physics and graphics engines derived from the video game industry, most notably the Unreal Engine \cite{dosovitskiy2017carla, rong2020lgsvl}. Simulators such as CARLA \cite{dosovitskiy2017carla} leverage these engines to provide high-fidelity urban environments and generate rich synthetic sensor data, including LiDAR, RGB, depth, and radar streams. These sensors are critical for training and validating perception, sensor fusion, and decision-making modules \cite{chitta2022transfuser, 10610842, 10919538}. 

\subsection{Scenario Generation Frameworks} 

Several tools have been developed to support scenario generation in AD, each varying in complexity and user requirements. For instance, Scenic \cite{fremont2019scenic} is a domain-specific probabilistic programming language that enables the concise specification of traffic scenarios with randomness and constraints. However, its custom syntax and programming requirements limit accessibility for non-expert users.

OpenSCENARIO \cite{openscenario2020} is a standard developed by ASAM for describing dynamic elements in a driving scenario, such as trajectories and events, using XML. While powerful and simulator-compatible, it is verbose and typically requires manual scripting. To simplify usage, OpenScenarioEditor \cite{openscenarioeditor2023} provides a graphical interface for composing OpenSCENARIO-compliant scenarios. Although it reduces the need to edit raw XML, it still assumes familiarity with the OpenSCENARIO schema and terminology, offers limited validation or feedback when composing complex behaviors, and lacks integrated simulation or execution support. %Scenarios built in the editor typically require additional setup using external tools like ScenarioRunner for use in environments such as CARLA, which must be configured separately—creating a fragmented workflow and increasing technical overhead.

RoadRunner \cite{roadrunner}, developed by MathWorks, provides an intuitive framework for creating road networks and traffic scenarios. While easy to use for basic design tasks, advanced functionality depends on MATLAB scripting, making it language-specific and less accessible to different user groups. Moreover, RoadRunner is a proprietary tool and not open source, which limits extensibility and integration with open research pipelines.

In summary, current scenario generation frameworks rely heavily on programming knowledge, are tied to specific languages or modeling standards, and are designed with simulation fidelity, not deep learning integration, in mind.

% CommonRoad, a framework from TU Munich, defines standardized benchmarks for motion planning, offering pre-built scenarios, evaluation tools, and a Python API. While it facilitates reproducible research and scenario variation, generating new scenarios from scratch or modifying existing ones requires programming skills and a good understanding of the object structure, which may not suit users seeking a low-code or ML-oriented interface \cite{althoff2017commonroad}.

\subsection{Advancements in Deep Learning Approaches} 
%INTEGRATE papers from https://arxiv.org/abs/2305.13960 Recent advancements in deep learning, such as RC-GANs and Recurrent Autoencoders, have improved scenario generation by modeling diverse driving trajectories \cite{2661-8907}. These methods facilitate realistic test scenarios for validation, as seen in frameworks that integrate trajectory clustering and anomaly detection. Additionally, deep learning-based heterogeneous driver models enable dynamic, stochastic scenario generation, reflecting diverse human driving behaviors \cite{s23094570}. However, there are also uncertainties related to deep learning, which can be classified into aleatoric uncertainty (noise and randomness in the data) and epistemic uncertainty (incomplete knowledge of the models) \cite{SUK2024317}. Therefore, expert-designed test cases remain essential, capturing rare or common-sense failures that automated methods may overlook. A hybrid approach that integrates deep learning with expert knowledge—in particular Knowledge-Guided Methods—is increasingly being adopted to accelerate learning and facilitate more comprehensive, robust validation of autonomous vehicles \cite{9512524}\cite{von2021informed}\cite{zhong2023guided}.

The rapid progress in deep learning has significantly advanced the field of AD, particularly through the emergence of end-to-end driving systems that learn control policies directly from raw sensory inputs \cite{hu2023planning, chitta2022transfuser}. To ensure the safety and reliability of such systems, there is a need for scenario generation tools that allow domain experts and policymakers to design and evaluate a wide range of complex and safety-critical driving scenarios. Furthermore, recent studies have proposed graph-based deep learning approaches \cite{feng2023trafficgen, suo2021trafficsim, mutsch2023model} for scenario generation, demonstrating their ability to create structured, diverse, and corner case scenarios crucial for rigorous validation. Tools that support standardized scenario formats compatible with these approaches enable seamless integration and facilitate experimentation within deep learning pipelines. 

Finally, behavior generation approaches \cite{rowe2024ctrl, rempe2022generating} integrated within simulation frameworks have been shown to significantly enhance the realism of agents' interactions, thereby improving the fidelity and realism of validation frameworks. These advancements collectively underscore the need for scenario generation tools that are not only modular and interoperable but also specifically designed to integrate seamlessly with modern learning-based scenario and behavior generation techniques.

%% file: sections/3_Methodology.tex
\section{Methodology}
% \textbf{What are we proposing from a very high level? (Sascha)} 
% At a high level, our framework provides a user-friendly graphical interface that allows users to load, create, modify, save and execute scenarios in one window without prior knowledge. A key component is our graph-based scenario representation, which simplifies scenario management and facilitates integration with deep learning methods for automated scenario generation. Furthermore, the framework supports random search-based scenario generation, ensuring diverse and comprehensive test cases for autonomous vehicle validation.

This work introduces a scenario generation framework designed to provide a user-friendly graphical interface that enables users to load, create, modify, save, and execute traffic scenarios—all within a single window and without requiring prior simulation knowledge or programming experience. Central to the framework is a graph-based scenario representation that streamlines scenario management and facilitates integration with deep learning techniques for automated scenario and behavior generation. 

\begin{figure}[!t]
    \centering
    \includegraphics[width=0.98\linewidth]{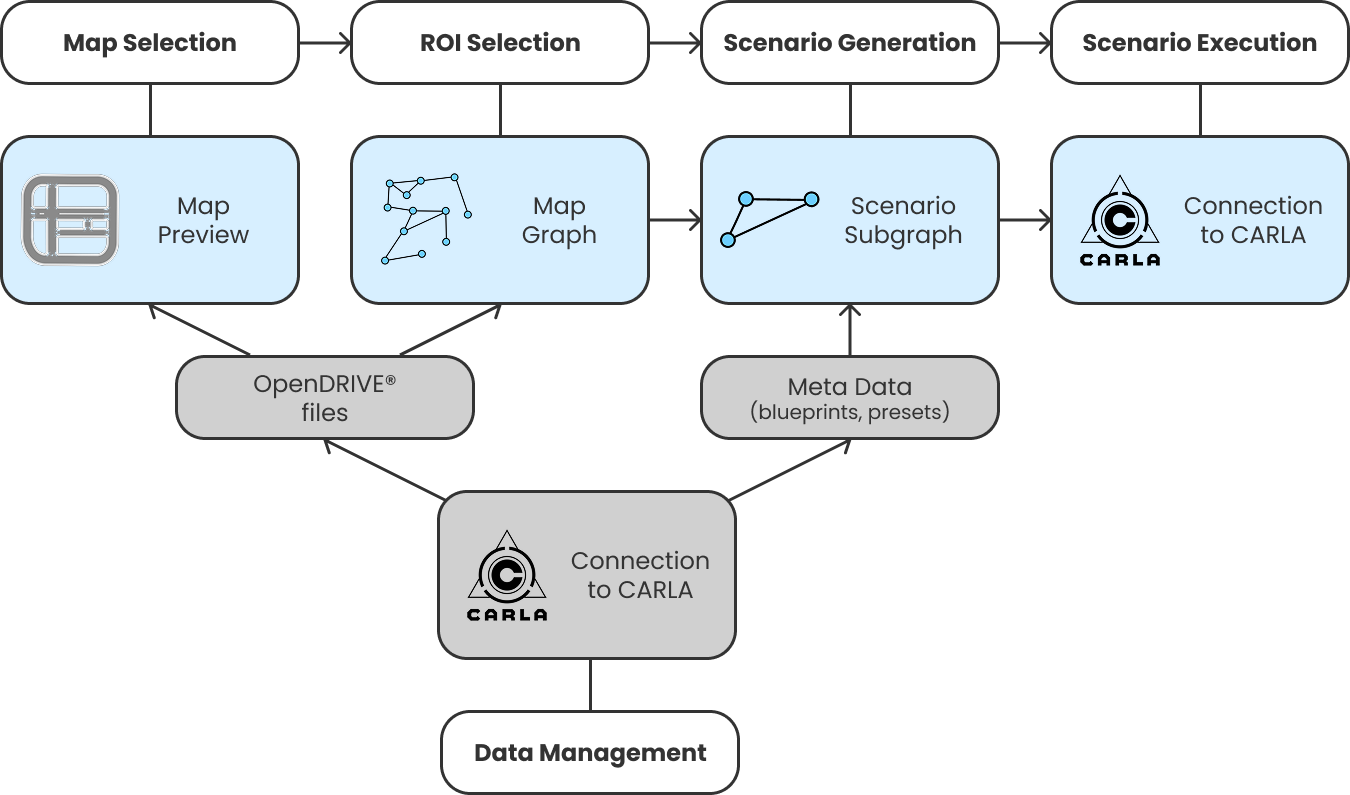}
    \caption{Overview of the proposed scenario generation workflow. The process begins with a one-time data management step, where the framework integrates a catalog of available OpenDRIVE maps and metadata, such as actor blueprints and environmental presets. Scenario creation then follows an intuitive flow: users select a map, define a region of interest (ROI), configure scenario parameters, and execute the scenario within the CARLA simulator.}
    \label{fig:diagramm}
\end{figure}

In addition, the framework supports random search-based scenario generation by assigning values to the nodes and edges of the proposed graph structure, promoting diversity and coverage in test cases for AVs validation with minimal manual input. Furthermore, it handles standard high-definition (HD) map formats, particularly the widely used OpenDRIVE format \cite{opendrive2005}, and is seamlessly integrated with the CARLA simulator \cite{dosovitskiy2017carla}. In the following sections, we present the key components and functionalities of the proposed framework, as illustrated in Fig. \ref{fig:diagramm}, highlighting how each contributes to a clear and intuitive scenario generation process.

% The framework is designed to guide users through a clear and intuitive scenario generation workflow. It begins by offering users a comprehensive catalog of supported maps, along with intuitive interaction tools and informative heuristics to guide scenario design. Once a map is selected, it is visually rendered in the interface, with key traffic elements such as crosswalks, stop signs, and traffic lights clearly highlighted to support realistic scenario creation. The map is then automatically divided into manageable segments, allowing users to flexibly define the spatial extent and duration of the scenario. After selecting one or more connected segments, users can populate the area with non-player characters (NPCs), configuring their behavioral parameters and navigation goals to simulate various traffic interactions. Once the scenario is complete, it can be saved and executed directly within the CARLA simulator, enabling immediate visualization and testing in a realistic virtual environment.
%%%%%%%%%%%%%%%%%%%%%%%%%%%%%%%%%%%%%%%%%%%%%%%%%%%%%%%%%
\subsection{Maps Integration and Data Management}
\label{map_integration}
A foundational element of the proposed framework is the use of HD road maps that enable users to create traffic scenarios of varying scales and complexities. Although multiple map formats exist, such as OpenStreetMap (OSM) \cite{haklay2008openstreetmap}, and Lanelet/Lanelet2 \cite{poggenhans2018lanelet2}, our framework is built around the OpenDRIVE format \cite{opendrive2005}. This preference arises from the fact that OpenDRIVE is developed specifically for high-fidelity driving simulations, offering detailed geometric descriptions of roads at the centimeter level, including elements such as curvature and elevation profiles that are crucial for accurate vehicle behavior modeling and safety validation.

In contrast to other formats that may require extensive post-processing or manual annotations, OpenDRIVE embeds critical traffic infrastructure elements, including lane connectivity and traffic lights, directly within the map structure. This built-in semantic richness eliminates the need for additional configuration and allows for immediate use in simulation. Furthermore, OpenDRIVE is widely supported across leading simulation platforms such as CARLA \cite{dosovitskiy2017carla} and LGSVL \cite{rong2020lgsvl}, making it a robust choice for scenario development.

The framework begins with an initialization phase, during which it processes available OpenDRIVE maps to both render them visually and extract semantic metadata, such as lane markings, crosswalks, speed limits, and road signs. These metadata are essential for enabling realistic scenario generation logic and provide transparency regarding the underlying traffic regulations. The framework also retrieves simulator-specific assets from CARLA, including the non-player character (NPC) blueprints and supported environmental conditions (e.g., weather presets). In addition to allowing users to update data when new assets become available, the framework provides a set of ready-to-use assets and maps to simplify and accelerate scenario development. This proposed data management flow offers a modular foundation that can be extended to other simulators, requiring only the provision of an OpenDRIVE map and a compatible interface for asset extraction and registration.

To optimize performance and reduce reliance on external processes, all extracted data is serialized into lightweight JSON files, allowing rapid access and parsing in subsequent steps. This design enables fully offline scenario generation, thereby removing the need for a constantly running simulator instance. As a result, users benefit from faster design interactions, improved system stability, and reduced resource consumption. 
%%%%%%%%%%%%%%%%%%%%%%%%%%%%%%%%%%%%%%%%%
%%%%%%%%%%%%%%%%%%%%%%%%%%%%%%%%%%%%%%%%%

\subsection{Map Selection} 
% with a focus on statistics and how they help with usability? figure for this page is required.(Erne)

% To enhance usability and streamline scenario creation, the map selection interface presents abstracted statistical metadata and a rendered visualization for each simulation environment, as illustrated in Fig. \ref{fig:map-selection}. By displaying this information upfront, users - particularly those unfamiliar with CARLA - can efficiently assess map complexity and suitability for specific testing goals.

% The selected characteristics reflect structural and functional aspects relevant to scenario diversity and validation coverage. Junction count serves as a proxy for navigational complexity, while crosswalk and traffic light metrics highlight potential for pedestrian interaction and traffic regulation testing. This abstraction helps users align map choice with intended behaviors, such as evaluating vehicle performance in high-conflict areas or under varying control regimes.

% This design reduces trial-and-error, lowers the entry barrier, and facilitates more targeted configuration. % It also enables automated and heuristic-based scenario generation  methods to leverage map metadata as filters or weighting criteria, enhancing the relevance and diversity of generated scenarios.

% The map selection page thus acts as a bridge between usability and technical functionality. It reduces cognitive overhead, promotes targeted test design, and aligns with the framework’s fundamental goal of making scenario generation both intuitive and scalable. 

\begin{figure}[!t]
    \centering
    \includegraphics[width=0.98\linewidth]{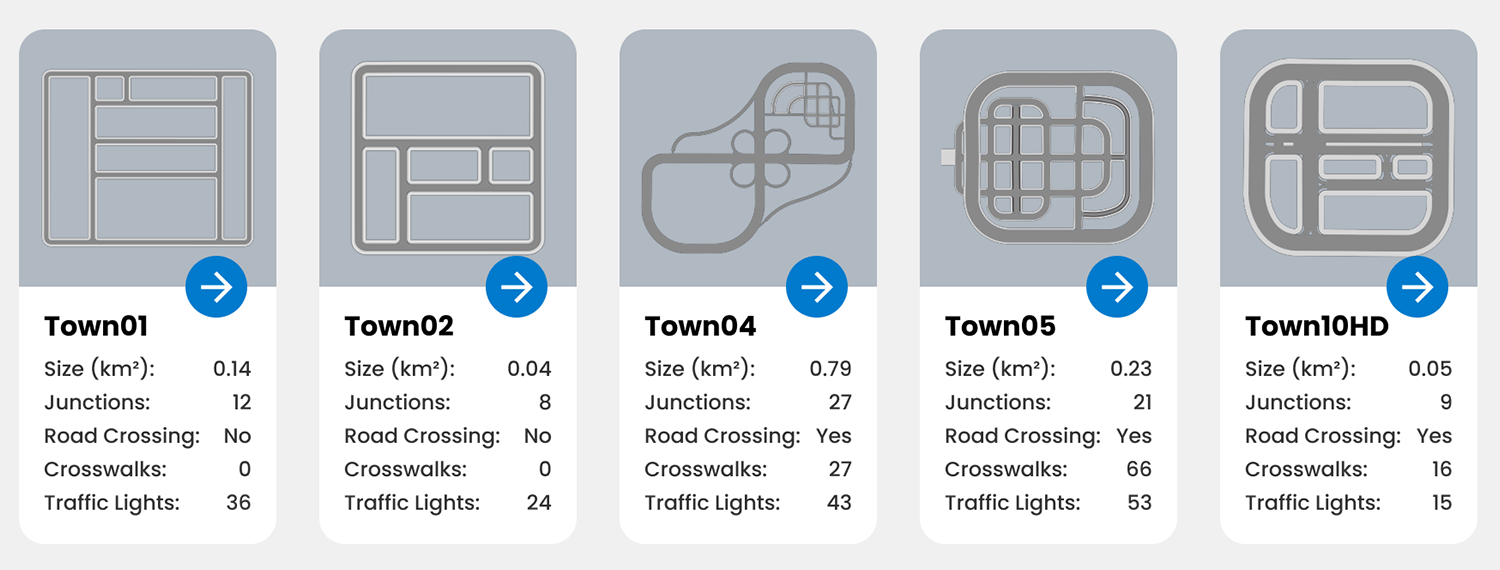}
    \caption{A catalog of OpenDrive maps extracted from CARLA, each annotated with statistical metadata, such as the number of junctions, crosswalks, and traffic lights, to help users identify a suitable map for scenario generation.}
    \label{fig:map-selection}
\end{figure}

To improve usability and simplify scenario setup, the map selection interface provides users with both abstracted statistical metadata and a rendered visualization of each map, as illustrated in Fig. \ref{fig:map-selection}. By presenting this information upfront, users can efficiently evaluate map complexity and its suitability for specific testing objectives.

The selected metadata highlights structural and functional attributes relevant to scenario diversity and validation coverage. Map size offers an estimate of spatial scale and the potential complexity of generated scenarios. The number of junctions serves as an indicator of navigational complexity, while metrics such as crosswalks and traffic lights reflect the potential for pedestrian interaction and compliance with traffic rules validation. This abstraction enables users to select maps that align with intended testing behaviors, such as assessing vehicle performance in high-conflict zones or under diverse traffic conditions.

Therefore, the map selection interface connects user-friendly interaction with underlying technical complexity. It reduces cognitive load, enables users to focus on use-case-specific scenario design, and supports the framework’s overarching objective of making scenario generation both flexible and intuitive.
%%%%%%%%%%%%%%%%%%%%%%%%%%%%%%%%%%%%%%%%%
%%%%%%%%%%%%%%%%%%%%%%%%%%%%%%%%%%%%%%%%%%%

\subsection{Region of Interest Selection} 
\label{roi}

Once a map is selected for scenario generation, users can define specific regions of interest (ROIs) within the map. This region selection mechanism enables scenario customization tailored to distinct research or application goals. For instance, short-term scenario generation typically targets smaller, localized regions involving a few actors, focusing on safety-critical interactions such as near-collisions, merges, or abrupt pedestrian crossings. These are often used to stress-test planning and control systems~\cite{hanselmann2022king,feng2023trafficgen}. In contrast, long-term scenario generation involves larger areas with denser traffic scenes. This mode is commonly adopted in end-to-end autonomy stacks to evaluate route planning, perception consistency, and behavior diversity over time~\cite{caesar2021nuplan, li2022metadrive}. By allowing users to define ROIs, our framework accommodates localized critical behaviors and extended spatiotemporal diversity in scenarios.

We propose a map representation that supports segmentation into ROIs, facilitating subsequent scenario generation. Although rasterized images are frequently employed to represent driving environments~\cite{chitta2024sledge}, the dense nature of images and sensitivity to rotation present challenges for scenario generation, particularly in terms of identifying feasible actor placements and necessitating extensive masking procedures. To address these limitations, we propose a graph-based representation of OpenDrive maps. Graphs offer a sparse and interpretable structure that not only simplifies the placement of traffic participants but also preserves essential road topology via relational edges, such as successor, predecessor, left, and right relations~\cite{liang2020learning}. Moreover, this graph formulation is suitable for integration with deep learning methods for trajectory prediction~\cite{vivekanandan2024ki}, scenario generation~\cite{feng2023trafficgen}, and clustering~\cite{zipfl2022clustering}. 

\begin{figure}[!t]
    \centering
    \includegraphics[width=0.75\linewidth]{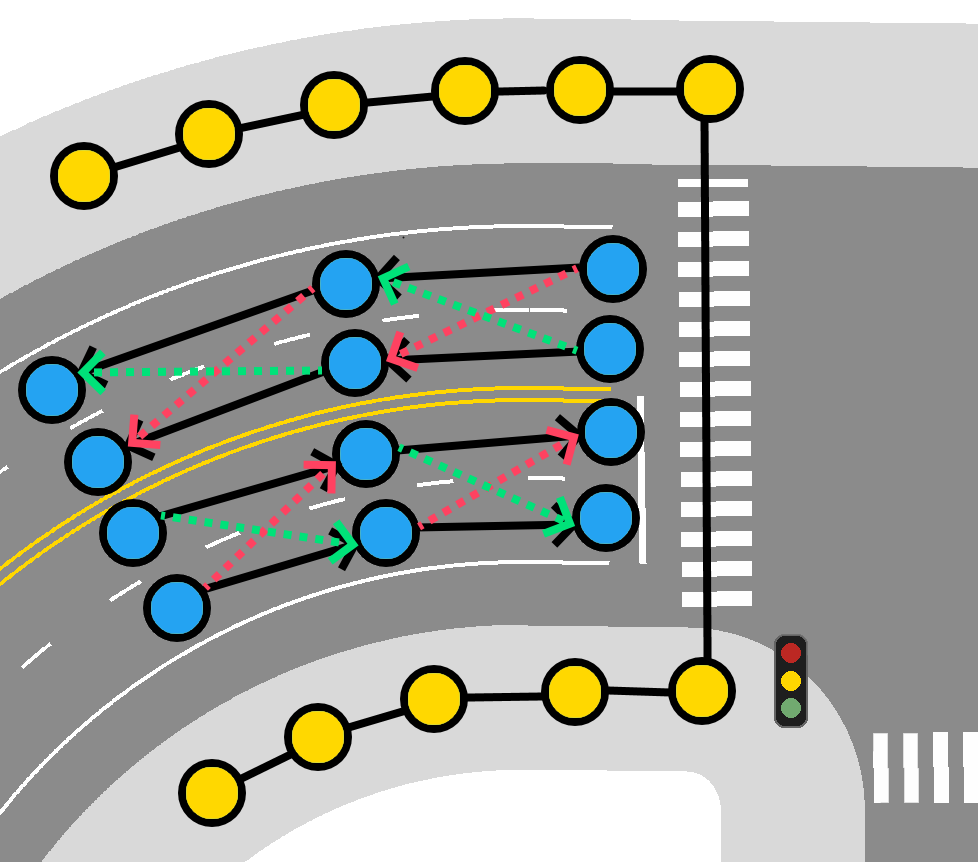}
    \vskip 1em \noindent
    {\footnotesize
    \begin{tabular}{lll}
        \textcolor{black}{\rule{1em}{1em}} \textbf{Successor Edge} & 
        \textcolor[HTML]{00E279}{\rule{1em}{1em}} \textbf{Right Edge} & 
        \textcolor[HTML]{FF4261}{\rule{1em}{1em}} \textbf{Left Edge} \\[0.25em] 
        
        \textcolor[HTML]{24A3F2}{\rule{1em}{1em}} \textbf{Road-bound Node} & 
        \textcolor[HTML]{FFD800}{\rule{1em}{1em}} \textbf{Non-Road-bound Node} 
        %\multicolumn{3}{c}{\textcolor[HTML]{24A3F2}{\rule{1em}{1em}} \textbf{Vehicle node} \hspace{2em}\textcolor[HTML]{FFD800}{\rule{1em}{1em}} \textbf{Pedestrian node}} 
    \\
    \end{tabular}
    }
    \caption{Visualization of the graph-based representation for a simplified map segment. Blue nodes indicate potential spawn locations for road-bound actors, while yellow nodes represent possible pedestrian spawn points. Directed edges—black for successor connections, green for right, and red for left—define the spatial relations between road-bound actor nodes. In contrast, undirected edges are used for pedestrians to capture their flexible movement.}
    \label{fig:graph-visual}
\end{figure}

Specifically, we represent the map as a directed graph $\mathcal{G = \{N, E}\}$, where \( \mathcal{N} \) denotes the set of nodes, interpreted as valid spawn points for NPCs, and \( \mathcal{E}\) represents the set of edges capturing the road topology. The nodes are uniformly sampled at equidistant intervals along the drivable lanes. Fig.~\ref{fig:graph-visual} shows such a graph in a simplified setting, focusing on a small section of the map for visual clarity. To facilitate this representation, the graph is constructed during the data management phase, discussed in section \ref{map_integration}, and stored in the GraphML format~\cite{brandes2002graphml}.

%%%%%%%%%%%%%%%%%%%%%%%%%%%%%%%%%%%%%%%%%%%
%%%%%%%%%%%%%%%%%%%%%%%%%%%%%%%%%%%%%%%%%%%
To support a structured and intuitive scenario generation process, the map graph is segmented into ROIs (i.e., subgraphs) based on the road topology defined by the OpenDRIVE specification. Each intersection is represented as a distinct subgraph, while roads are divided into one or more subgraphs depending on their length, ensuring approximately uniform segment sizes. The segmentation process maintains the connectivity information between adjacent subgraphs, preserving the topological structure of the original map and enabling seamless composition of larger scenarios.

\begin{figure}[!t]
    \centering
    \begin{subfigure}[t]{0.48\linewidth}
        \centering
        \includegraphics[width=\linewidth]{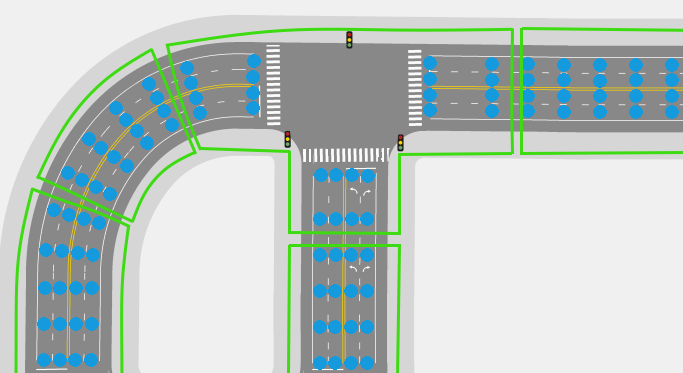}
        \caption{The user is prompted to select a base ROI, which defines the initial region for scenario generation.}
        \label{fig:subfig-roi}
    \end{subfigure}%
    \hfill
    \begin{subfigure}[t]{0.48\linewidth}
        \centering
        \includegraphics[width=\linewidth]{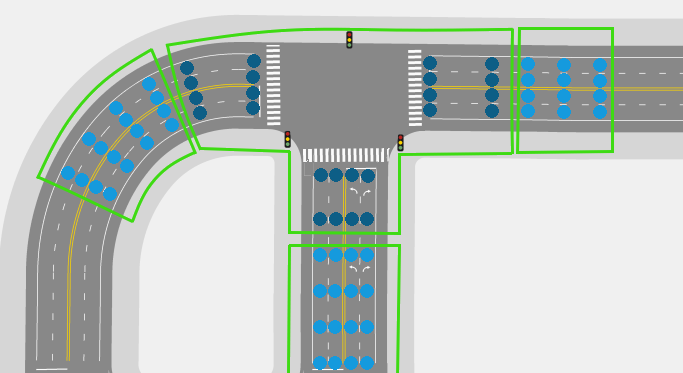}
        \caption{After the user selects the base ROI (shown in dark blue), adjacent regions eligible for extension are highlighted in light blue.}
        \label{fig:subfig-scenario}
    \end{subfigure}
    \caption{Interactive process for defining the ROI used in scenario generation. The interface guides users in selecting an initial ROI and expanding it through adjacent, connected regions, ensuring spatial continuity in the overall scenario ROI.}
    \label{fig:area-selection}
\end{figure}

%Building on this segmentation, users can interactively select a specific subgraph as the initial region for scenario generation, as illustrated in Fig.~\ref{fig:subfig-roi}. Additional subgraphs can be included to expand the selected region, with the constraint that only connected subgraphs may be added, as clarified in Fig.~\ref{fig:subfig-scenario}. This ensures that the resulting overall scenario ROI is connected.
Building on this segmentation, users can interactively select a specific subgraph to serve as the initial region for scenario generation, as illustrated in Fig.~\ref{fig:subfig-roi}. This initial selection acts as the foundation for defining the scenario's ROI. The expansion of this ROI follows a strictly incremental process: at each step, users can add additional subgraphs, but only if they are directly connected to any subgraph already included in the ROI. This means that once a new subgraph is added, all of its immediate neighbors, along with those of previously selected subgraphs, become eligible for selection, as clarified in Fig.~\ref{fig:subfig-scenario}. This iterative mechanism ensures that the final scenario ROI forms a coherent area within the map, preserving logical consistency for agent interactions and traffic flow.

After the user selects the overall scenario ROI, it is further enriched with an undirected pedestrian graph. In this graph, nodes are sampled along the roadside to represent plausible pedestrian spawn points situated near the drivable areas. Edges are incorporated based on the spatial layout of sidewalks and crosswalks, capturing feasible pedestrian movements and interactions. The use of an undirected graph captures the flexibility of pedestrian motion, which is generally not governed by strict directional rules. This structure enables the modeling of realistic pedestrian behavior, which is essential for generating meaningful and safety-critical traffic scenarios. An example of the generated pedestrian graph is shown in Fig.~\ref{fig:graph-visual}.

%%%%%%%%%%%%%%%%%%%%%%%%%%%%%%%%%%%%%%%%%%%
%%%%%%%%%%%%%%%%%%%%%%%%%%%%%%%%%%%%%%%%%%%
%%%%%%%%%%%%%%%%%%%%%%%%%%%%%%%%%%%%%%%%%%%%%%%%%%
\subsection{Scenario Generation} 
\label{scene_gen}

\begin{figure*} [!t]
    \centering
    \begin{subfigure}[t]{0.32\linewidth}
        \includegraphics[width=\linewidth]
        {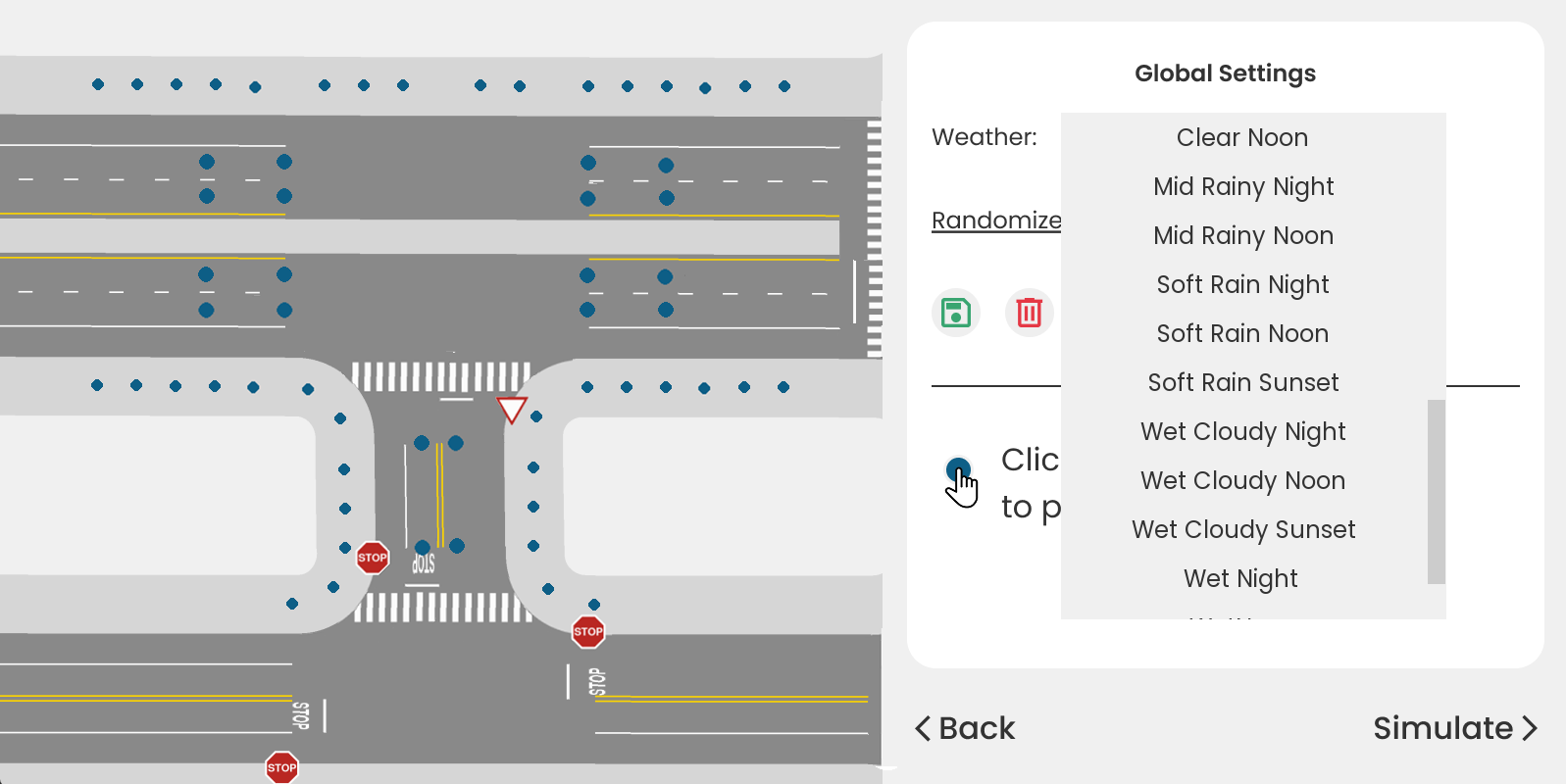}
        \caption{Interface for configuring weather and time of day, enabling users to select from predefined presets representing diverse environmental conditions.}
        \label{fig:sub-weather}
    \end{subfigure}
    \hfill
    \begin{subfigure}[t]{0.32\linewidth}
        \includegraphics[width=\linewidth]{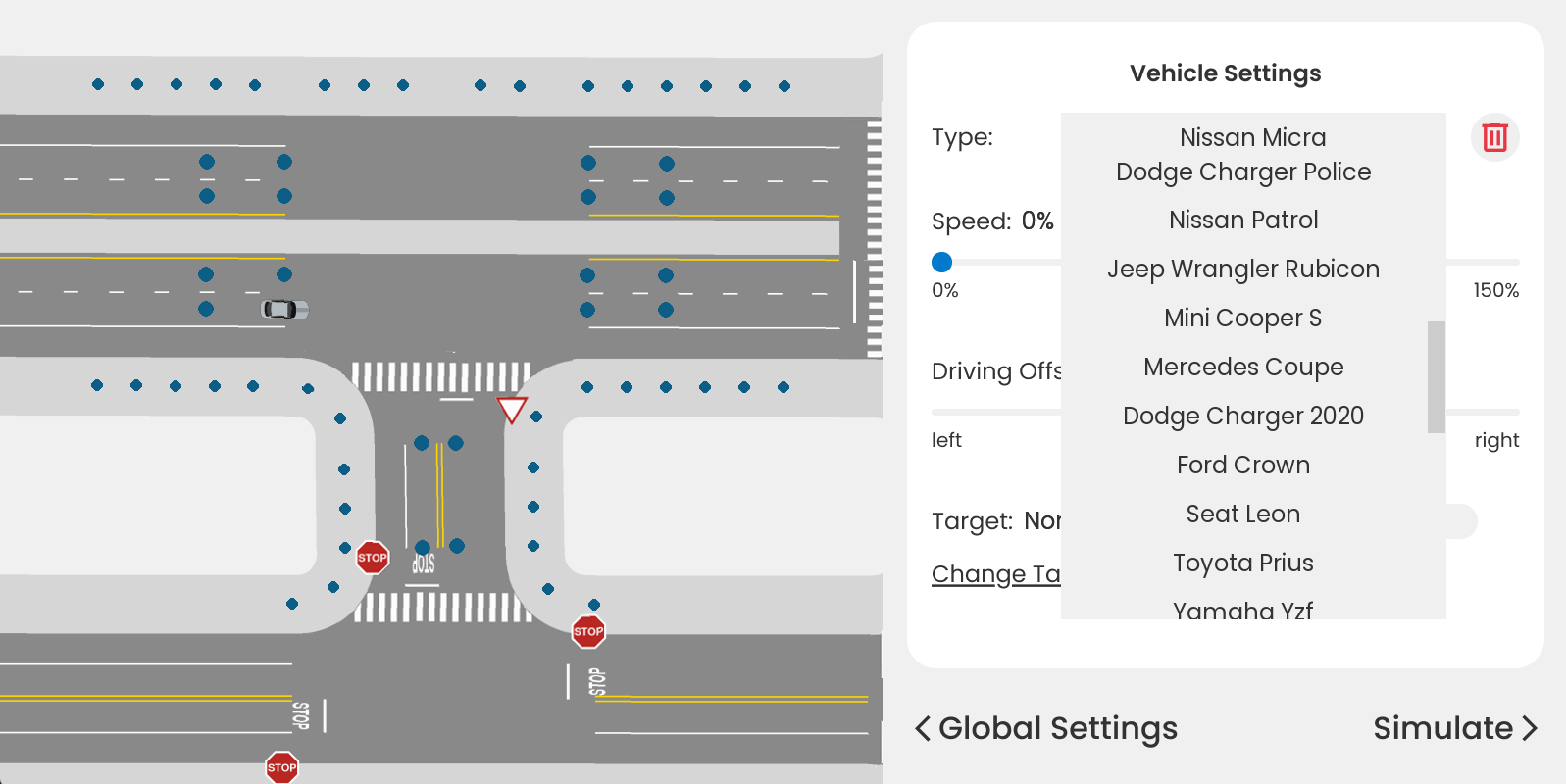}
        \caption{The NPC configuration interface allows the user to begin by selecting a valid spawn point, then choose an NPC category and specific model to spawn.}
        \label{fig:sub-car}
    \end{subfigure}
    \hfill
    \begin{subfigure}[t]{0.32\linewidth}
        \includegraphics[width=\linewidth]{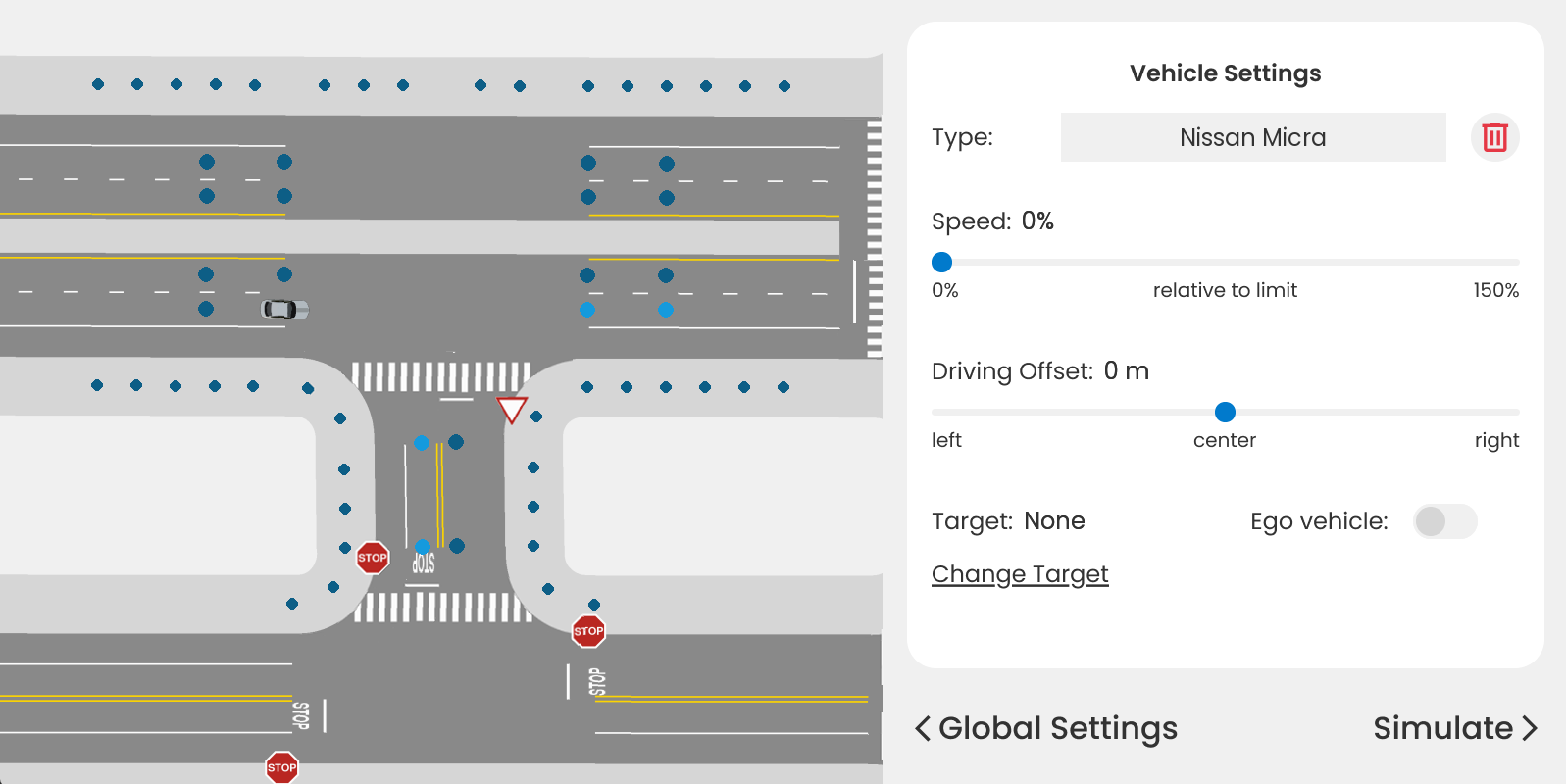}
        \caption{The target assignment interface highlights reachable goal nodes (in light blue) based on the selected spawn point. The user selects a target suited to their use case and configures the NPC’s dynamic parameters.}
        \label{fig:sub-target}
    \end{subfigure}
        \caption{User-guided configuration workflow for scenario generation. The interface supports sequential selection of environmental conditions, NPC placement, goal assignment, and behavioral parameter configuration.}
        \label{fig:empty-scenario}
\end{figure*}

Following the selection of relevant ROI within the map, the induced subgraph corresponding to the selected region is utilized for scenario generation. Scenario generation begins with the user configuring environmental parameters such as weather conditions and time of day, as illustrated in Fig.~\ref{fig:sub-weather}. These options include simulator-supported settings like rain, fog, and nighttime. These environmental factors are critical for testing autonomous systems under conditions that can considerably influence both vehicle dynamics and perception systems performance \cite{tian2018deeptest}.

To support a broad spectrum of validation conditions, our proposed framework enables the direct integration of various actor types into the scenario, thereby enhancing diversity in scenario design. During the data management phase, described in section \ref{map_integration}, all available actors from CARLA are retrieved and systematically categorized into seven semantic groups: normal vehicle, pedestrian, bicycle, motorcycle, van, truck, and bus. These categories are visually differentiated within the user interface, improving usability by allowing users to easily recognize and manage different actor types during scenario composition. In addition to semantic grouping, users are also given the option to select specific models from the asset catalog, as demonstrated in Fig.~\ref{fig:sub-car}, providing greater flexibility and control over scenario customization.
%%%%%%%%%%%%%%%%%%%%%%%%%%%%%%%%%%%%

The placement of actors is facilitated through an interactive process, where users select valid spawn nodes from the induced subgraph, as detailed in section~\ref{roi}. Once a spawn point is chosen, the actor’s trajectory is defined through a reachability analysis, which identifies all candidate goal nodes. Goal candidates are defined as terminal nodes—those without outgoing left, right, or successor edges—representing the endpoints of navigable paths, as illustrated in Fig.~\ref{fig:sub-target}. From these candidates, the user can select a goal node that aligns with the intended behavior they wish to simulate, such as merging, crossing an intersection, or navigating through a specific segment. 

Users can further customize actor behavior by adjusting dynamic parameters such as desired velocity and driving offset, as illustrated in Fig.~\ref{fig:sub-target}. This level of configurability allows the creation of nuanced behaviors and the exploration of edge cases. For instance, a cyclist riding slightly off-center in the lane ahead of the ego vehicle may introduce a challenging overtaking scenario. In contrast, a slow-moving vehicle could disturb the overall traffic flow and increase accident risk. By enabling control over the motion parameters, the framework avoids generating overly idealized scenarios in which agents exhibit perfectly consistent behavior. Such realism is essential to prevent overfitting during autonomous system validation, as real-world traffic often includes imperfect lane keeping and variable driving speeds~\cite{wang2024survey}.

Furthermore, incorporating vulnerable road users (VRUs), such as pedestrians, is essential for comprehensive scenario design. Their inherently unpredictable behavior, particularly in situations involving individual or group interactions, introduces critical test cases that challenge the effectiveness of perception and planning modules in detecting and responding to VRUs~\cite{silva2024vulnerable}. Within our framework, pedestrians are integrated similarly to other actors, utilizing the pedestrian undirected subgraph for spawn node selection, goal assignment, and walking speed configuration. 

Furthermore, the framework allows users to designate any vehicle within the scenario as the ego vehicle, representing the autonomous system under test. Depending on the scenario layout and the configured interactions, the ego vehicle can be placed in situations that validate its behavior in critical driving contexts, such as overtaking, yielding, emergency braking, or navigating complex intersections.

To ensure a compact and efficient representation of generated scenarios, actor-specific parameters—including actor category (or asset type), driving speed, and driving lateral offset—are encoded as attributes of the designated spawn node within the scenario subgraph. An edge is also added between the spawn node and the actor’s goal node—either specified by the user or randomly selected if not provided—which serves as a reference for trajectory planning during scenario realization.

Additionally, the framework incorporates an optional fully automated mode designed to generate a diverse dataset of scenarios without requiring user input. In this mode, maps and ROIs are selected through random sampling, followed by the construction of scenarios based on probabilistic selection of key parameters. These parameters include the scenario fill percentage, defined as the ratio of spawned actors to the maximum allowable number of actors, the spawn nodes of actors, their categories, behavioral parameters, and corresponding goal nodes. Furthermore, the framework supports integration with deep learning-based scenario generation methods, such as those proposed in \cite{suo2021trafficsim, feng2023trafficgen}, by explicitly providing empty scenario subgraphs as input. These methods can then generate complete scenarios by modifying the subgraph, adding actors, and specifying their dynamic behaviors.
%%%%%%%%%%%%%%%%%%%%%%%%%%%%%%%%%%
% \subsection{Scenario Realization} 
%Using CARLA, traffic manager, and Walker AI. 
%Report that we can highlight the ego vehicle for validation of different driving software.
%the traffic behavior can be generated with other algorithms.
%2 examples of scenarios generated using the tool (left: scenario in framework, right: realization of scenario in CARLA).

%% file: sections/5_evaluation.tex
\section{Evaluation}
% Once a scenario has been configured, the framework enables its execution within CARLA using the built-in Traffic Manager and Walker AI modules. This phase translates the abstract scenario definition into a dynamic simulation, where all elements—vehicles, pedestrians, traffic flow-are instantiated and controlled according to the specified parameters.
% A key feature of the realization process is the ability to stream live simulation footage from a bird’s-eye perspective directly into the framework. This overhead view provides immediate visual feedback, bridging the gap between high-level scenario specification and low-level behavioral observation. The real-time stream allows users to rapidly assess whether agent behavior aligns with intended configurations and identify any discrepancies without leaving the application environment. This tight coupling between scenario design and execution improves usability and enables more efficient debugging and validation.
% To illustrate this capability, Fig. \ref{fig:filled-scenario} and Fig. \ref{fig:carla_simulation} present two examples of scenarios generated. Each example shows the scenario as configured within the framework (Fig. \ref{fig:filled-scenario}) and its execution within the CARLA simulator (Fig. \ref{fig:carla_simulation}), demonstrating visual consistency and behavioral alignment between design and deployment stages.

This section presents the evaluation of the proposed framework, with a particular emphasis on scenario realization—namely, how configured scenarios are executed within the CARLA simulator. Once a scenario is defined, the framework leverages the associated scenario subgraph to spawn actors in the simulated environment and assign their dynamic behaviors and goal positions. During runtime, actors are controlled using CARLA's built-in Traffic Manager and Walker AI modules, with their trajectories generated based on the shortest path between their spawn and goal locations.

A user-friendly feature of the realization process is the live streaming of simulation footage from a bird’s-eye perspective directly within the framework interface. This bird’s-eye view provides immediate visual feedback, bridging the gap between high-level scenario configuration and low-level behavioral observation. The real-time stream enables users to assess whether agent behavior aligns with the intended setup and identify deviations without leaving the application environment. This close integration of scenario design and execution improves usability and supports more efficient debugging and validation. After the simulation of a generated scenario is completed, the framework seamlessly returns the user to the scenario generation interface, discussed in section \ref{scene_gen}, enabling further editing and refinement. This iterative process supports granular scenario validation and enhances the user experience through a streamlined workflow.

\begin{figure}[!t]
    \centering
    \begin{subfigure}[b]{0.98\linewidth}
        \centering
        \includegraphics[width=\linewidth]{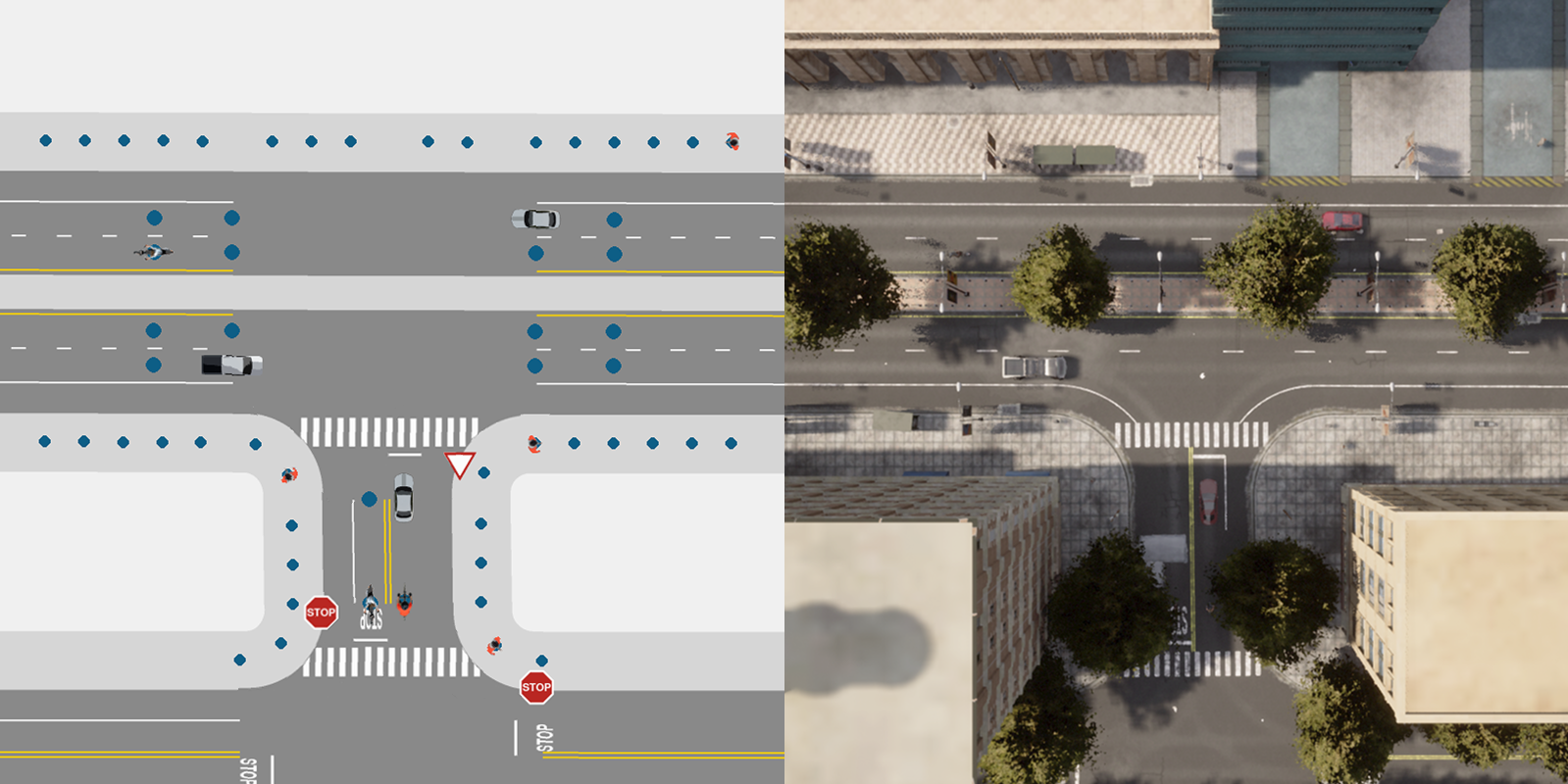}
        \caption{Daytime scenario configured in the framework (left) and executed in CARLA (right), demonstrating preserved road layout, agent behavior, and traffic logic.}
        \label{fig:carla_simulation_a}
    \end{subfigure}
    \vskip\baselineskip
    \begin{subfigure}[b]{0.98\linewidth}
        \centering
        \includegraphics[width=\linewidth]{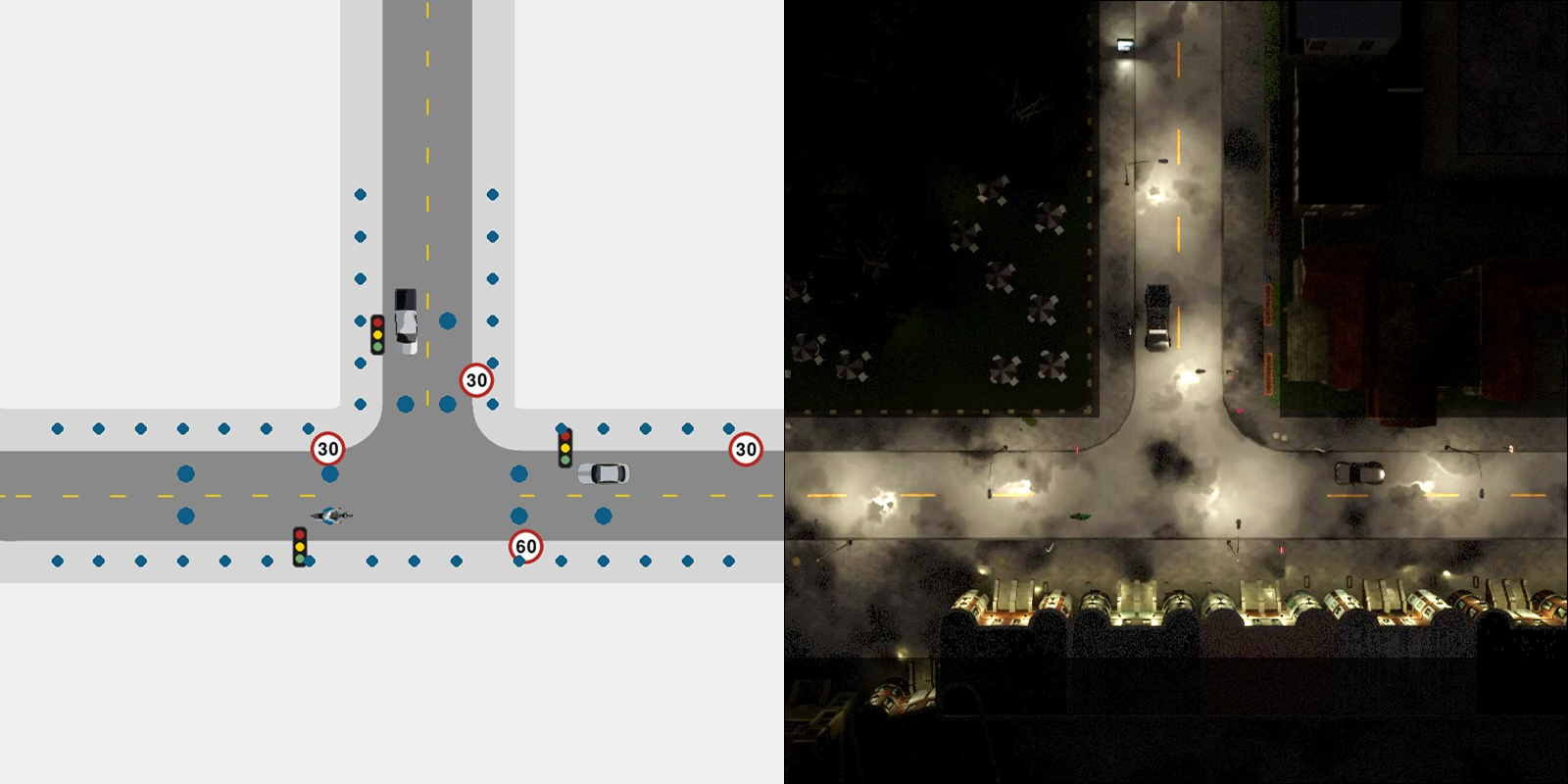}
        \caption{Nighttime scenario, simulated in CARLA under low-light conditions, maintaining consistent agent behavior and scene setup.}
        \label{fig:carla_simulation_b}
    \end{subfigure}
    \caption{Two example scenarios designed using the framework and executed in CARLA. Each pair shows the scenario configuration and its corresponding simulation, demonstrating visual and behavioral fidelity across different lighting conditions.}
    \label{fig:carla_simulation_combined}
\end{figure}

To demonstrate the framework’s ability to translate designed scenarios into simulated executions, Fig. \ref{fig:carla_simulation_a} and Fig. \ref{fig:carla_simulation_b} showcase two example scenarios. Each figure illustrates the configuration of a scenario within the framework and its execution in the CARLA simulator, highlighting the visual and behavioral consistency between design and execution stages.
%%%%%%%%%%%%%%%%%%%%%%%%%%%%%%%%%%%%%%%%%%%%
%%%%%%%%%%%%%%%%%%%%%%%%%%%%%%%%%%%%%%%%%%%%

%% file: sections/6_conclusion.tex
\section{Conclusion}
In this paper, we presented a user-friendly, no-code framework for scenario generation, designed to simplify and democratize scenario-based validation for AVs. The framework allows users to graphically create, modify, and run complex scenarios without programming, using a graph-based representation that supports both manual and automated scenario generation. The framework’s ability to automatically sample scenario parameters, such as actor types, behaviors, and environmental conditions, helps generate diverse and realistic test datasets. Real-time visualization and built-in CARLA modules facilitate efficient debugging and validation. By lowering technical barriers, the framework empowers a wider range of stakeholders in autonomous system testing. Future work will focus on integrating real-world traffic data and incorporating heuristic or learning-based methods to enhance scenario diversity and realism.